# Dynamics-informed deconvolutional neural networks for super-resolution identification of regime changes in epidemiological time series

Jose M. G. Vilar[1,2,*] and Leonor Saiz[3,*]

[1] Biofisika Institute (CSIC, UPV/EHU), University of the Basque Country (UPV/EHU), P.O. Box 644, 48080 Bilbao, Spain

[2] IKERBASQUE, Basque Foundation for Science, 48011 Bilbao, Spain

[3] Department of Biomedical Engineering, University of California, 451 E. Health Sciences Drive, Davis, CA 95616, USA

[*]Correspondence to: j.vilar@ikerbasque.org or lsaiz@ucdavis.edu

## Abstract

Inferring the timing and amplitude of perturbations in epidemiological systems from their stochastically spread low-resolution outcomes is as relevant as challenging. It is a requirement for current approaches to overcome the need to know the details of the perturbations to proceed with the analyses. However, the general problem of connecting epidemiological curves with the underlying incidence lacks the highly effective methodology present in other inverse problems, such as super-resolution and dehazing from computer vision. Here, we develop an unsupervised physics-informed convolutional neural network approach in reverse to connect death records with incidence that allows the identification of regime changes at single-day resolution. Applied to COVID-19 data with proper regularization and model-selection criteria, the approach can identify the implementation and removal of lockdowns and other nonpharmaceutical interventions with 0.93-day accuracy over the time span of a year.

## Introduction

Inferring the underlying causative signal from the observed effector response through an extended system in space, time, or both is a fundamental problem in many fields. The key challenge is to obtain the input signal as a function of the system output. This type of inverse problem is widespread in multiple disciplines, and in many of them, such as computer vision, has reached highly sophisticated levels[1,2]. Dehazing, for instance, aims at recovering the scene radiance as if there were no perturbations along the line of sight from the natural environment, including fog, dust, and other low-visibility weather[3]. In single-molecule localization microscopy, super-resolution beyond the width (~250 nm) of the point spread function (PSF) is achieved by deconvolving the visualized light field to infer the precise position (with an uncertainty down to ~10 nm) of the light emitters[4]. Deconvolution of the image field is also the key element of deblurring, the process of going back from an image blurred through a distortion, such as an out-of-focus projection or objective motion, to the original, sharper image[5]. Blurring is typically modeled as the convolution of the sharper image with the blur kernel, characterized by a PSF as well, that leads to the output image and defines the inverse problem. Compounded with all these effects, there is noise, which degrades the output and brings about the need of using regularization methods[6].

Inverse problems in epidemiological systems involve inferring the daily incidence, defined as new infections per day, from the number of infection-caused deaths or positive testing individuals. New daily infections over time provide the input signal. As the disease progresses in an infected individual, it has a probability of being detected as well as a probability of leading to death at every time point after infection. Each of these probabilities depends on the characteristics of the disease and is population specific, including age structure, clinical resources, and testing capabilities[7]. Daily positive testing results and daily deaths provide the measured outputs. This problem in epidemiology compounds multiple problems like those present in computer vision. As in blurring, the output of daily new infections is spread, in this case over time instead of over space; as in hazing, the output provides distorted information because not all infections are detected or lead to death; and, as in super-resolution, it needs to provide information at a much finer level than the spread of the output. In addition, the output is also inherently noisy due to, among others, the fundamental stochasticity of the testing and death processes. Because of all these effects compounded together and the traditional widespread lack of reliable field data availability, the methodology in epidemiological systems has not been as developed as in computer vision and other fields.

The state of the art in the field is epitomized by the deconvolution of daily cases or mortality time series to infer infection incidence. For instance, death records have been used to infer the influenza incidence through an iterative method with early stopping as a regularization approach, which prevents the amplification of noise originating from the inherent randomness of the death process[8]. Deconvolution of infection cases has also been used to reconstruct the infection curves for SARS epidemics[9] through Expectation Maximization Smoothing (EMS) method[10]. In the case of COVID-19, there have also been recent refinements to account for model misspecification and censored observations[11] and to estimate new infections in real-time[12]. The prototypical state-of-the-art approach provides much more reliable information than simply considering incidence as the number of newly detected positive cases[13] but is not sufficiently reliable to provide a daily incidence from which to infer discontinuities in transmission, as determined by jumps in the instantaneous reproduction number, which will change immediately following the implementation of a nonpharmaceutical intervention (NPI)[14].

Here, we develop a convolutional neural network approach in reverse to connect death records with incidence that recovers the underlying dynamics with such a precision as to allow the identification of regime changes at single-day resolution. We address major challenges that in computer vision are not as exacerbated, including the presence of a high dynamic range (three orders of magnitude, ranging from zero to over a thousand deaths per day); increasing and decreasing exponential growth regimes (doubling times and half-lives of ~3 days); kernels widely spread over time (mean of ~20 days with ~10-day standard deviation (SD)); and the presence of large amounts of noise (Poisson processes with mean below one). To this end, we derive an objective function to be optimized through the network that considers a suitable loss for the underlying Poisson process, is quasi-scale invariant, and includes, following a physics-informed neural network (PINN) approach, the determinants of the epidemiological dynamics. The PINN contribution to the objective function promotes sparsity in the changes in disease transmission as defined by the underlying renewal equation for the incidence. Methodologically, the approach faces the challenges of deblurring with components of dehazing, denoising, and super-resolution.

To benchmark the approach with field data, we focus on the effects of major NPIs, including lockdowns, on the global spread of the COVID-19 outbreak[15]. Being able to identify, with single-day resolution, the implementation time of major NPIs from the epidemiological data has not been achieved so far. It is important from the biomedical point of view to test for potential differences in the clinical evolution of the disease over time and over different countries[16]. Existing analysis of restrictions, imposed quarantines, lockdowns, and other NPIs have estimated a substantially spread delay between the implementation of lockdowns and their effects on the dynamics of the outbreak as measured by the basic reproduction number from case counts[17]. However, this delay and spread should not be present in the infection transmission. Other approaches based on compartmental models[18] or Bayesian analyses[19] have relied on the inclusion of temporal-specific information, namely, the date of the NPI, to estimate the amplitude of the effects[20]. From a fundamental point of view, the COVID-19 outbreak provides a unique opportunity to contrast the methodology with diverse types of field data. The detailed characterization over small, controlled populations has provided detailed information from the infection to death or recovery and the chain of infections. This characterization involved testing for the causative virus SARS-CoV-2, identifying contacts and the infection time, and following up the clinical evolution[7,21-23]. Therefore, it is highly relevant to verify to what extent this characterization holds at a large scale (country-wide level) over diverse heterogeneous populations (different countries).

## Results

### Convolutional approach

We have derived a physics-informed convolutional neural network able to identify and quantify the distinct growth regimes of the epidemiological dynamics as described below.

Mathematically, new infections (incidence) and deaths on day $t$ are described as Poisson processes with intensities $i_t$ for the number of daily new infections and $\lambda_t$ for the number of daily deaths. The intensities of these two processes are related to each other through the convolution

$$\lambda_t = \sum_{\tau=0}^{t-1} f_\tau r_{t-\tau} i_{t-\tau} = \sum_{\tau=0}^{t-1} f_\tau j_{t-\tau},$$

(1)

where $f_\tau$ denotes the time from-infection-to-death probability mass function, which gives the probability for a death to have occurred on day $\tau$ after infection, and $r_t$ refers to the infection fatality ratio (IFR), which gives the probability for an individual infected on day $t$ to die according to $f_\tau$. We express the convolution in terms of the scaled incidence $j_t = r_t i_t$, which represents the number of new infections on day $t$ that will eventually result in death. Throughout the derivations below, we consider the time interval $t \in [1, T]$.

### Scale-invariant probabilistic loss function

The number of daily deaths $n_t$ is assumed to follow a Poisson distribution with parameter $\lambda_t$,

$$P_{n_t} = \frac{\lambda_t^{n_t}}{n_t!} e^{-\lambda_t},$$

(2)

which implies a log-likelihood function for $\lambda_t$ given by

$$l(n_t, \lambda_t) = n_t \ln \lambda_t - \lambda_t - \ln(n_t!).$$

(3)

Therefore, the intensity of the Poisson process $\lambda_t$ corresponds to the expected number of daily deaths.

There are multiple loss functions that can potentially be used. The focus is on growth rate changes, namely, relative changes, which are invariant upon scaling of the population by a constant factor. To account for this invariance in the loss function, which using the Stirling approximation can be rewritten as $l(n_t, \lambda_t) \simeq n_t \ln \frac{\lambda_t}{n_t} + n_t - \lambda_t$, we weight the negative log-likelihood function at each time point by $\lambda_t^{-1}$, leading to

$$\mathcal{L}_\Lambda = -\frac{1}{Z} \sum_{t=1}^{T} \frac{1}{1+\lambda_t} l(n_t, \lambda_t), \quad \text{with } Z = \sum_{t=1}^{T} \frac{1}{1+\lambda_t},$$

(4)

as a loss function for the overall time interval. This loss function has a minimum value for $\lambda_t = n_t, \forall t \in [1, T]$. The term 1 is added to the denominator to avoid undetermined results for sustained time intervals with zero deaths.

**Dynamics and regularization**

The epidemiological dynamics is characterized by the renewal equation $i(t) = \int_0^\infty k_I(t, \tau) i(t - \tau) d\tau$, with $k_I(t, \tau)$ being the rate of secondary transmissions per single primary case[24,25]. This convolution equation provides the basis for the definitions of the instantaneous reproduction number $R_t = \int_0^\infty k_I(t, \tau) d\tau$ and the probability density of the generation time $w(\tau) = \frac{k_I(t,\tau)}{R_t}$. The instantaneous reproduction number $R_t$ accounts for the transmission of the infectious population at time $t$ and the generation time $w(\tau)$ describes the infectiousness profile at time $\tau$ after infection.

We consider the discrete counterpart of the dynamic equations and changes in the infection fatality ratio at a slower time scale than the incidence so that the discrete renewal equation can be rewritten in terms of the normalized incidence as

$$j_t = R_t \sum_{\tau=1}^{t-1} w_\tau j_{t-\tau},$$

(5)

where $w_\tau$ is the probability mass function of the generation time.

We implement a regime-change-aware regularization approach through

$$\mathcal{L}_R = \frac{1}{T-1} \sum_{t=1}^{T-1} |\ln R_{t+1} - \ln R_t|$$

(6)

as regularization term in the objective function. This expression implies sparseness in $|\ln R_{t+1} - \ln R_t|$ and therefore it is suitable to identify discontinuous changes in $R_t$. The logarithm is used to account for expanding and decaying dynamics symmetrically.

**Dynamics-informed convolutional neural network**

Analogously to the stablished physics-informed neural network approach, we initially consider the loss function of the epidemiological dynamics as $\mathcal{L}_D = \frac{1}{T-1}\sum_{t=2}^{T}(j_t - R_t \sum_{\tau=1}^{t-1} w_\tau j_{t-\tau})^2$. Strictly enforcing the epidemiological dynamics, $\mathcal{L}_D = 0$, leads to $R_t = j_t / \sum_{\tau=1}^{t-1} w_\tau j_{t-\tau}$, which allows the combination of $\mathcal{L}_D$ and $\mathcal{L}_R$ into a unified loss

$$\mathcal{L}_{RD} = \frac{1}{T-2} \sum_{t=2}^{T-1} \left| \ln \frac{j_{t+1}}{j_t} - \ln \frac{\sum_{\tau=1}^{t-1} w_\tau j_{t-\tau}}{\sum_{\tau=1}^{t-1} w_\tau j_{t+1-\tau}} \right|,$$

(7)

which we use to include both the dynamics and regularization in the objective function to be optimized by the neural network.

Explicitly, we infer the estimated scaled incidence $\hat{\mathbf{J}} = (\hat{j}_1, \ldots, \hat{j}_T)$ as

$$\hat{\mathbf{J}} = \underset{j_t: \forall t \in [1,T]}{\mathrm{argmin}} \, (\mathcal{L}_\Lambda + \gamma \mathcal{L}_{RD}),$$

(8)

where $\gamma$ sets the strength of the regularization contribution. The structure of the neural network is shown in Fig. 1. It is based on a dual convolution neural network in reverse. Namely, the direct approach infers the convolution kernels from input and output signals. Instead, we infer the unknown signal (scaled incidence) from the known kernels (time from-infection-to-death and generation time probability mass functions) and output signal (daily deaths). Therefore, we use as input layer the kernel with values $f_\tau$, which is fed into a transposed convolution layer with weights $j_t$, which is both connected to the output layer with values $\lambda_t$ and regularized according to the dynamics through a convolution with the kernel $w_\tau$.

**Model selection**

Through model selection we infer the optimal strength of the regularization term $\gamma$. We follow explicitly two approaches.

Prospectively, $\gamma$ is selected according to the Akaike information criterion[26] as the value of $\gamma$ that leads to the maximum of

$$L_{AIC} = \sum_{locations} \left( 2 \sum_{t=2}^{T-1} \Theta(|\ln R_{t+1} - \ln R_t|) - 2T\mathcal{L}_\Lambda \right),$$

(9)

where $\Theta(\cdot)$ is the Heaviside step function. The term $\sum_{t=2}^{T} \Theta(|\ln R_{t+1} - \ln R_t|)$ represents the number of parameters in the model, as shown generally for the number of nonzero parameters in lasso problems[27], and $T\mathcal{L}_\Lambda$ is the maximum value of the loglikelihood function for the model, expressed as the daily average loglikelihood times the number of days.

Retrospectively, $\gamma$ is selected as the value that minimizes the overall mean squared error between inferred and actual NPI implementation times for all the locations.

**Application of the approach reveals a discontinuous incidence**

We applied the approach to the identification of regime changes in the COVID-19 infectious dynamics in European locations and to the quantification of how these changes correlate with the timing of NPIs. Explicitly, the expected number of daily deaths $\lambda_t$ and the scaled incidence $j_t$ were obtained by training the network over the daily deaths with the generation time $w_\tau$ and infection-to-death kernels $f_\tau$ as inputs (Fig. 1). The focus is on locations with the exact date of death on record to target the precision of the approach at the single-day level, which is substantially smaller than the observed >3-day average delay in reporting deaths[28,29], and on the first year of the outbreak, the period that concentrated major NPIs before the widespread evolution of viral variants, substantial vaccination, and large levels of acquired immunity.

The results (Fig. 2), after model selection with the strength of the regularization term $\gamma = 2.51$, show the daily deaths fluctuating around their inferred expected values $\lambda_t$ without any substantial bias, with higher relative fluctuations at low expected values. The overall average value of the relative fluctuations, quantified as $\langle (n_t - \lambda_t)^2 / \lambda_t \rangle$, is 1.25 deaths/day. The fact that this value is essentially 1 confirms the ability of the network to consistently track the nature of the underlying Poisson process over a wide dynamic range, represented on a logarithmic scale, from values below 1 up to above $10^3$, with both expanding and decaying dynamics.

In stark contrast to the smoothness of the expected number of deaths $\lambda_t$, the network leads to a rugged, discontinuous behavior for the scaled incidence $j_t$. These discontinuities, as implied by the infection transmission described by Eq. (5), originate from a discontinuous behavior in the instantaneous reproduction number. The key question is if the incidence recovered from the smooth fit to daily deaths accurately describes the epidemiological dynamics. In general, the inferred changes in the instantaneous reproduction number can correspond to the regime changes or to optimal segmentations of a continuously changing dynamics. The ability of the approach to capture the effects of NPIs, which are expected to lead to regime changes, provides a stringent validation avenue.

**Validation identifies regime changes with major NPIs**

Explicitly, the validation of the approach is conducted by predicting the precise timing of major NPIs, including the implementation and removal of lockdowns and other restrictions in social interactions (Supplementary Table S1). The date of the NPI is inferred as the date of the largest $R_t$ change in a 9-day window centered at the date of a recorded NPI. We define the offset, $\Delta$, as the inferred minus the actual time of the NPI.

For the initial stages of the outbreak, the approach accurately identifies the dates of the major lockdowns and social restrictions, with an overall offset for all locations of 0.22 (mean) ± 0.63 (SD) days (Fig. 3). These results are particularly remarkable both in their precision, considering that incidence stochastically spreads into deaths with a delay of 19.3 days on average and width of ±9.1 (SD) days, and in their uniformity, with all the parameters of the network being the same for all the locations, including the value of the regularization term. For the whole period considered, the approach retained the accuracy at the single-day level, with an overall offset of -0.07 ± 0.92 days (Fig. 4). In addition to the high concordance across countries uncovered in the initial stages of the outbreak, identification of NPI timings over a year also highlights the consistency over time, as no systematic biases are observed.

The case of England, with an overall offset of 0.19 ± 0.93 days, is particularly relevant to illustrate the temporal invariance of the clinical and epidemiological dynamics, as all the major discontinuous changes correspond to major NPIs (Fig. 4). Besides the clinical parameters of the disease remaining statistically invariant over time, a major reason for this excellent agreement is the number of daily deaths remaining always at high values, with the lowest value of $\lambda_t$ never crossing 10 deaths/day (Fig. 2). In contrast, the approach missed the progressive lifting of NPIs from May to September in countries with extremely low deaths (Switzerland, and Denmark), which reached values of $\lambda_t$ below 0.2 deaths/day (Fig. 2). The quality of the data also plays a significant role. In the case of Italy, there are major changes in June-August for $R_t$ inferred from deaths in July-September with similar overall values to those of England but with an unusual distribution, dichotomously alternating randomly between zero and values around 10 deaths/day. Concomitantly, there was no support to potentially associate the resulting changes in $R_t$ for Italy with obvious NPIs during this period.

**Super-resolution identification of lifting and implementing NPIs**

Our results also show that there are no major mechanistic differences between implementing and lifting NPIs. As a clear example, lifting of the second England nationwide lockdown in December 2020 and implementation of the third one in January 2021 were both inferred to happen within one day of their actual date (Fig. 4 and Supplementary Table S1). In this case, lifting and implementing an NPI happen at a similarly high incidence, which is not generally the case. Systematic statistical differences may arise, in general, because low incidence parallels high fluctuations and lifting NPIs sides with low incidence, which makes it more challenging to infer properties of the dynamics[30]. This difference is also important when concurrent measures are applied because lockdowns affect mostly regions with high incidence and would be more noticeable than releasing restrictions in low-incidence regions, which might be hidden by the high-incidence regions.

Altogether, the distribution of the offset of NPI inferred times is much narrower, about an order of magnitude narrower, than the distribution of infection-to-death delay used in the inference process (Fig. 5). Such a precision constates that the approach presented here provides a super-resolution identification of NPI timings comparable to the general state of the art in super-resolution localization microscopy[4], and in the same range as in biomolecular imaging in cells[31]. Here, the date of the NPI is identified as the actual date, the day before, or the day after in 86% of the cases (Fig. 5).

The epidemiological dynamics plays a fundamental role through the strength of the regularization term. The value of $\gamma = 2.51$ corresponds to the retrospective model selection approach and provides the optimal constraint. This value can be estimated prospectively through the Akaike information criterion. In this case, it is essential to have the proper probabilistic model for the data. Excluding Italy, which strongly deviates from the Poisson death statistics in July-September, model selection led to a similar value $\gamma = 2.00$, which resulted in an overall offset of 0.00 ± 1.10 days. This concordance indicates that the unsupervised approach can generally identify regime changes at the single-day resolution.

## Discussion

We have derived a dynamics-informed deconvolutional neural network (DIDNN) to infer the underlying epidemiological dynamics at single-day resolution from death records. The main computational challenge has been to transfer the information in highly fluctuating daily deaths backward in time to new infections and forward again to infection transmissibility changes through two kernels up to the identification of regime changes at the single-day level. The kernels have a spread as large as a 9.1-day standard deviation with a 19.3-day average delay. In this regard, the increase in resolution of the prediction is a factor 10x over the information available at the daily death level. A key aspect of the success of the approach has been the incorporation of the epidemiological dynamics with proper regularization into the network through a PINN-based methodology[32,33]. This development allowed the approach to infer discontinuities in the instantaneous reproduction number to account for NPIs. At a fundamental level, our results show that an apparently smooth death dynamics does not result from a similarly continuous evolution of new infections but from a discontinuous, non-differentiable incidence dynamics. Additional aspects were the consideration of the proper fluctuation model (Poisson process) and scale invariance (dynamics governed by relative changes, not absolute changes) in the objective function.

Previous approaches to identifying the effects of NPIs relied primarily on tracking the case-based reproduction number over time. In these instances, it has been argued that the effects of NPIs are delayed from 7 days[34,35] up to 3 weeks[17,36] for the implementation and an extra week[17], up to a total of 4 weeks[36] for the lifting. In part, these delays arise from considering incidence as the number of newly reported cases rather than new infections, which introduces at least a delay of a week from the average time from infection to symptom onset. This delay is compounded with the time lag from symptom onset to reporting. On the methodological side, there are also delays introduced by averaging in real-time, both by non-centered moving averages over the data and by the inherent averaging procedure used to compute the instantaneous reproduction number[37]. Overall, all these processes lead to delayed blurring of the actual incidence. The advantages of using case counts are that their numbers are substantially larger than deaths, therefore reducing the noise, and that the average delay from infection to illness is shorter than to death, therefore allowing for an easier mathematical estimation without performing the needed deconvolutions. The disadvantages are the potential lack of testing capabilities, which were extreme in the early stages of the COVID-19 pandemic, and the still undetermined nature of the delays. Uncertainties in the case-based approach become even greater in a contact-tracing context, as asymptomatic cases can typically be detected up to 4 weeks after infection[38]. Therefore, case-based approaches are generally not suitable for the time period we considered nor to target resolutions at the single-day level.

Our results, in contrast, show that the effects of major NPI generally happen instantaneously in the disease transmission, as generally expected[14] and as assumed in previous approaches that relied on death counts to infer the effects of NPIs with known implementation dates[19].

From the epidemiological side, our results show that the same kernel for the infection-to-death time distribution can accurately account for the timing of major NPIs over multiple countries over the time span of a year. This consistency implies the same properties for the underlying clinical features of the disease over time and over similar populations (European countries that recorded death dates). In this regard, the kernel we used is the same as the one used in previous studies[19] with the consideration that deaths occur on average 3.6 days earlier than their reporting.

From the computational point of view, our results have paved the way for the inclusion of fully-integral dynamic equations into PINNs, which have traditionally been restricted to systems with differential time evolution, such as PDEs, ODEs, and integrodifferential equations[39]. More broadly, our approach has provided an avenue to generally implement complex model-aware regularization protocols in deconvolution methodologies[40] and to explicitly use the dynamics of the system as a part of the regularization process.

## Methods

### Field data

The data was downloaded from the "The Demography of COVID-19 Deaths" curated database[41] except for England. The UK reports by default deaths within 28 days of a positive test, which misses a substantial part of deaths. For England we considered daily "deaths within 60 days of a positive test by death date"[42], which provides a more stringent compromise between deaths by and deaths with COVID-19.

Data for NPIs was obtained from online resource[43] and NPI dataset[20]. Multiple consecutive NPIs were assigned the middle-point date rounded to days.

### Kernel parametrization

The kernels are parametrized according to the available clinical and field data and are based on previous parametrizations[19].

The generation time kernel, $w_\tau$, follows a gamma distribution with a mean of 6.3 days and a standard deviation (SD) of 4.2 days[19]. This parametrization is the same as for the serial interval[44] and is consistent also with the characterization using UK household data, which led to similar average (5.9 days, 5.2–7.0 days) and SD (4.8 days, 4.0–6.3 days)[45].

The time from-infection-to-death kernel, $f_\tau$, has been estimated to approximately follow a gamma distribution with 22.9-day mean and 9.1-day SD for reported deaths[19]. Since we use actual death dates, we shortened the mean to 19.3 days to account for the reporting delay[28,29]. The values selected are consistent with UK field data estimates for the time from-infection-to-death distributions, with average values in the range from 17.4 to 24.7 days[46].

### Computational Implementation

The overall approach was implemented in Keras[47] with custom losses coded in TensorFlow[48]. Logarithms of quantities susceptible of being zero were calculated with $10^{-1}$ added to them to avoid divergences during the training process. For each country, we considered 41 independent values of the regularization term, equally spaced on a logarithmic scale, from $\gamma = 10^{-1}$ to $\gamma = 10$. In model selection, the quantity $|\ln R_{t+1} - \ln R_t|$ was considered to be zero for values below $10^{-3}$.

Because sparse regularization is non-local, the initial growing trend was propagated backwards in time to avoid the effects of the widespread loss of the death counts (lack of diagnosis) during the very early stages of the outbreak. The cutoff time to initiate backwards propagation was set for each country as the last day the cumulative number of deaths was smaller than 1% of the maximum daily deaths.

## Acknowledgments

J.M.G.V. acknowledges support from Ministerio de Ciencia e Innovación under grants PGC2018-101282-B-I00 and PID2021-128850NB-I00 (MCI/AEI/FEDER, UE).

## Contributions

J.M.G.V. and L.S. designed research, performed research, analyzed data, and wrote the paper.

## Competing Interests

The authors declare no competing financial interests.

## Supplementary Information

Supplementary Table S1

# Figures

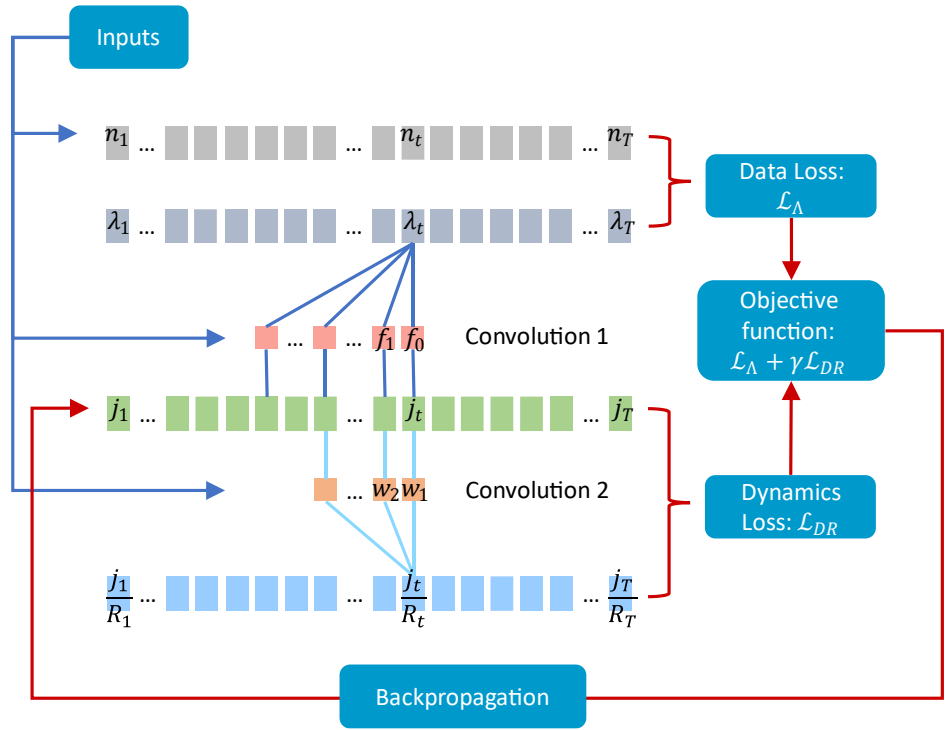

**Figure 1. Structure of the dynamics-informed deconvolutional neural network (DIDNN) for super-resolution identification of epidemiological regime changes.** A CNN is used in reverse. The inputs consist of the filters of the two convolutions, $w_\tau$ and $f_\tau$, and the daily deaths, $n_t$. The daily scaled incidence $j_t$ is obtained after minimization of the objective function through backpropagation, which includes both data, $\mathcal{L}_\Lambda$, and dynamics-informed, $\mathcal{L}_{DR}$, losses. The forward convolutions lead to the estimated expected daily deaths, $\lambda_t$, (convolution 1), and to the estimated instantaneous reproduction number, $R_t$ (convolution 2).

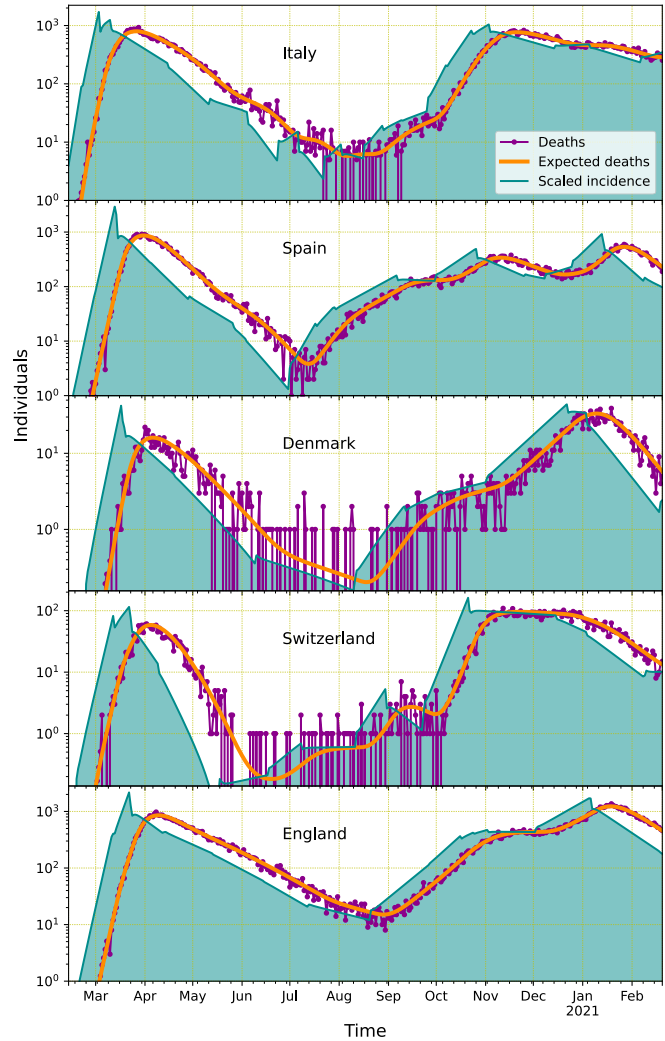

**Figure 2. The DIDNN infers an irregular, discontinuous incidence through the smoothed dynamics of the fluctuating daily deaths.** The scaled incidence (filled green curve) for the different locations follows from the corresponding number of daily deaths (magenta dotted line) through the smooth time courses of their expected value (orange lines).

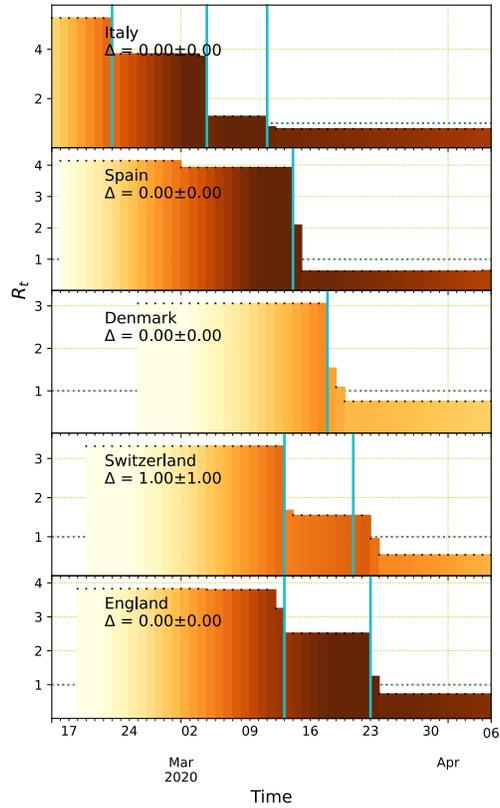

**Figure 3. Discontinuous changes in the instantaneous reproduction number are accurately associated with NPI timings.** The detail of the early-stage dynamics for the scaled incidence, $j_t$, as in Fig. 2 and color-coded in each panel from light (1 individual/day) to dark ($10^3$ individuals/day) tones, shows an underlying stepwise instantaneous reproduction number, $R_t$ (black dots). The offset, Δ, between NPI timings (cyan vertical lines) and major $R_t$ changes is indicated for each location in days as mean ± SD, with an overall offset for all locations of 0.22 ± 0.63 days.

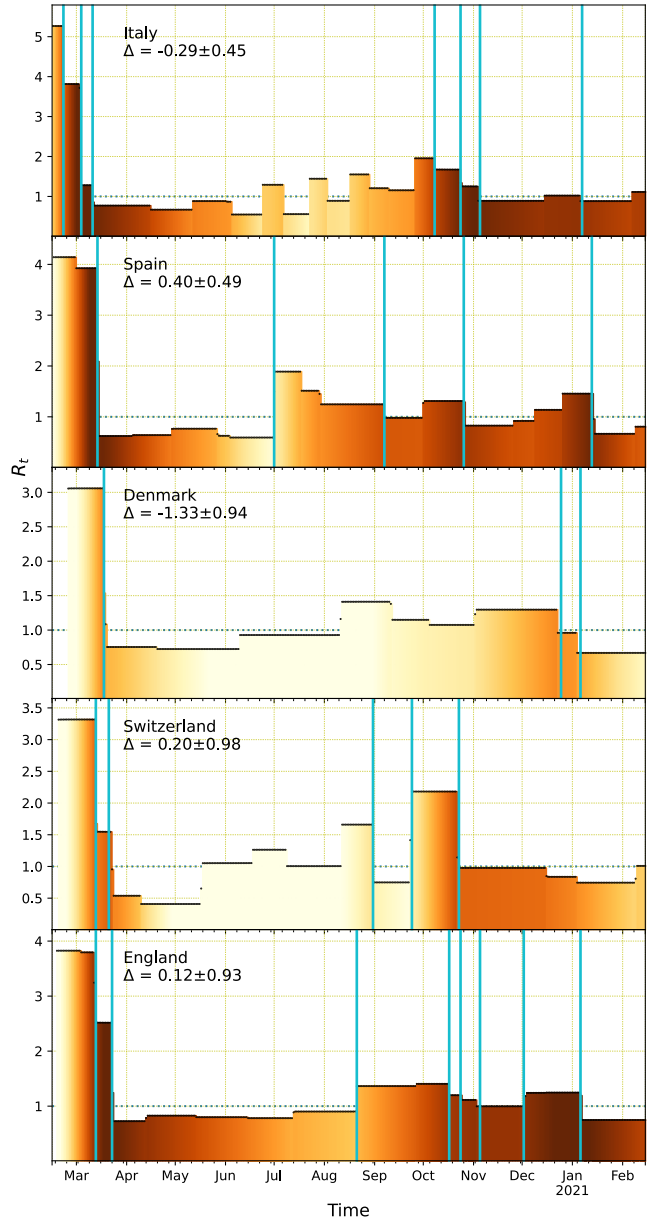

**Figure 4. The infection-to-death dynamics remains homogenous across locations and over time.** The time courses of $R_t$ and $j_t$ accurately track the major NPIs over a year with the same distribution of infection-to-death delay (average of 19.3 days and SD of 9.1 days). $R_t$, $j_t$, and NPI timings are represented as in Fig. 3. The offset, $\Delta$, between NPI timings and major $R_t$ changes are indicated for each location in days as mean ± SD, with an overall offset for all locations of -0.07 ± 0.92 days.

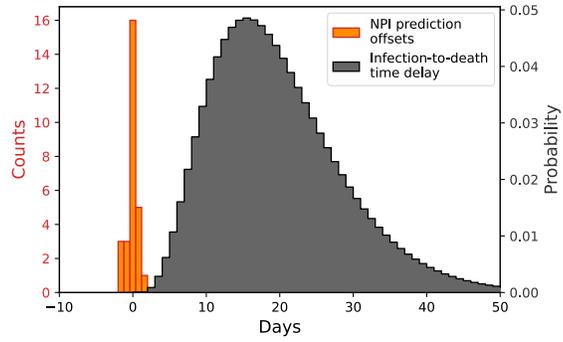

**Figure 5. Super-resolution identification of NPI timings.** The distribution of the offset of NPI inferred times (orange bars, left axis), with 0.92 SD, is much narrower than the distribution of infection-to-death delay used in the inference process (grey curve, right axis), with 9.1 SD.